\documentclass[twoside,11pt]{article}

\usepackage{jmlr2e}
\usepackage{caption}
\usepackage[toc,page]{appendix}
\usepackage{listings}
\usepackage{python}
\usepackage{tabularx}
\usepackage{xspace} 

\def\CC{{C\nolinebreak[4]\hspace{-.05em}\raisebox{.4ex}{\tiny\bf ++}}}

\newcommand{\tick}{\texttt{tick}\xspace}
\newcommand{\sklearn}{\texttt{scikit-learn}}
\newcommand{\lightning}{\texttt{lightning}}
\newcommand{\hawkesR}{\texttt{hawkes} \textsf{R}}
\newcommand{\ptpack}{\texttt{PtPack}}
\newcommand{\pyhawkes}{\texttt{pyhawkes}}

\ShortHeadings{\texttt{tick}: a Python library for statistical learning}{Bacry, Bompaire, Ga\"{\i}ffas and Poulsen}
\firstpageno{1}

\begin{document}

\title{\texttt{tick}: a Python library for statistical learning, with a particular emphasis on time-dependent modeling}

\author{\name Emmanuel Bacry \email emmanuel.bacry@polytechnique.edu \\
       \name Martin Bompaire \email martin.bompaire@polytechnique.edu \\
       \name St\'ephane Ga\"{\i}ffas \email stephane.gaiffas@polytechnique.edu \\
       \name S{\o}ren V. Poulsen \email soren.poulsen@polytechnique.edu \\
       \addr Centre de Math\'ematiques Appliqu\'ees\\
       \'Ecole polytechnique\\
       UMR 7641, 91128 Palaiseau, France}

\editor{xxxx}

\maketitle

\begin{abstract}%
\tick is a statistical learning library for Python~3, with a particular emphasis on time-dependent models, such as point processes, and tools for generalized linear models and survival analysis.
The core of the library is an optimization module providing model computational classes, solvers and proximal operators for regularization.
\tick relies on a \CC{} implementation and state-of-the-art optimization algorithms to provide very fast computations in a single node multi-core setting.
Source code and documentation can be downloaded from \url{https://github.com/X-DataInitiative/tick}.
\end{abstract}

\section{Introduction}

The aim of the \tick library is to propose to the Python community a large set of tools for statistical learning, previously not available in any framework.
Though \tick focuses on time-dependent modeling, it actually introduces a set of tools that allow to go way beyond this particular set of models, thanks to a highly modular optimization toolbox. It benefits from a thorough documentation (including tutorials with many examples), and a strongly tested API that brings to the scientific community cutting-edge algorithms with a high level of customization.
Optimization algorithms such as SVRG~\citep{johnson2013accelerating} or SDCA~\citep{shalev2013stochastic} are among the several optimization algorithms available in \tick that can be applied (in a modular way) to a large variety of models. 
An emphasis is done on time-dependent models: from the Cox regression model~\citep{andersen2012statistical}, a very popular model in survival analysis, to Hawkes processes, used in a wide range of applications such as geophysics~\citep{ogata1988statistical}, finance~\citep{bacry2015hawkes} and more recently social networks~\citep{zhou2013learning,xu2016learning}.
To the best of our knowledge, \tick is the most comprehensive library that deals with Hawkes processes, since it brings parametric and nonparametric estimators of theses models to a new accessibility level.

\section{Existing libraries}

\tick follows, whenever possible, the \sklearn{} API \citep{pedregosa2011scikit,buitinck2013api} which is well-known for its completeness and ease of use, 
 which makes it the reference Python machine learning library. 
However, while \sklearn{} targets a wide spectrum, \tick has a more specific objective: implementing highly-optimized algorithms with a particular emphasis on time-dependent modeling (not proposed in \sklearn{}). 
The \tick optimization toolbox relies on state-of-the-art optimization algorithms, and is implemented in a very modular way. 
It allows more possibilities than other \sklearn{} API based optimization libraries such as \lightning{}\footnote{\url{http://contrib.scikit-learn.org/lightning}}.

A wide variety of time-dependent models are taken care of by \tick, which makes it the most comprehensive library that deals with Hawkes processes for instance, by including the main inference algorithms from literature.
Despite the growing interest in Hawkes models, very few open source packages are available. 
There are mainly three of them.
The library \pyhawkes\footnote{\url{https://github.com/slinderman/pyhawkes}} proposes 
a small set of Bayesian inference algorithms for Hawkes process. 
\hawkesR{} \footnote{\url{https://cran.r-project.org/web/packages/hawkes/hawkes.pdf}} is a \textsf{R}-based library that provides a single estimation algorithm, and is hardly optimized. 
Finally, \ptpack{}\footnote{\url{https://github.com/dunan/MultiVariatePointProcess}} is a C++ library which proposes mainly parametric maximum likelihood estimators, with sparsity-inducing regularizations.
However, \ptpack{} is not interfaced with a user-friendly scripting language such as Python, which makes it less accessible to end-users for quick prototyping and experimenting on datasets. Moreover, as illustrated below, \ptpack{} exhibits poor performance compared to \tick.

\section{Package architecture}

The \tick library has four main modules: \texttt{tick.hawkes} for Hawkes processes (see Section~\ref{sec:hawkes} for a detailed review), \texttt{tick.linear\_model} with linear, logistic and Poisson regression, \texttt{tick.robust} for robust regression and \texttt{tick.survival} for survival analysis. 
Each of these modules provide simulation tools and learners to easily learn from data.
Whenever possible, \tick follows the \sklearn{} API.
The core of \tick is made of easy to combine penalization techniques (proximal operators), available in the \texttt{tick.prox} module and several convex solvers, available in the \texttt{tick.solver}, to train almost any available model in the library, see Table~\ref{tab:optim-class-examples}
for a non-exhaustive list of possible combinations.
An exhaustive list is available on the documentation web page\footnote{\url{https://x-datainitiative.github.io/tick/}}, and is given in Figure~\ref{fig:tick_structure} of the supplementary material.
 

\begin{figure}
  \centering
  \small
  \begin{tabular}{c c c}
  Model & Proximal operator & Solver \\
  \hline \hline
  Linear regression & L2 (Ridge) & Gradient Descent \\
  Logistic regression & L1 (Lasso) & Accelerated Gradient Descent \\
  Poisson regression & Total Variation& Stochastic Gradient Descent \\
  Cox regression & Group L1 & Stochastic Variance Reduced Gradient \\
  Hawkes with exp. kernels & SLOPE & Stochastic Dual Coordinate Ascent \\
  \end{tabular}
  \captionof{table}{\small \tick allows the user to combine many models, prox and solvers}
  \label{tab:optim-class-examples}
\end{figure}

\section{Hawkes}
\label{sec:hawkes}

\begin{figure}
  \centering
  \includegraphics[width=\textwidth]{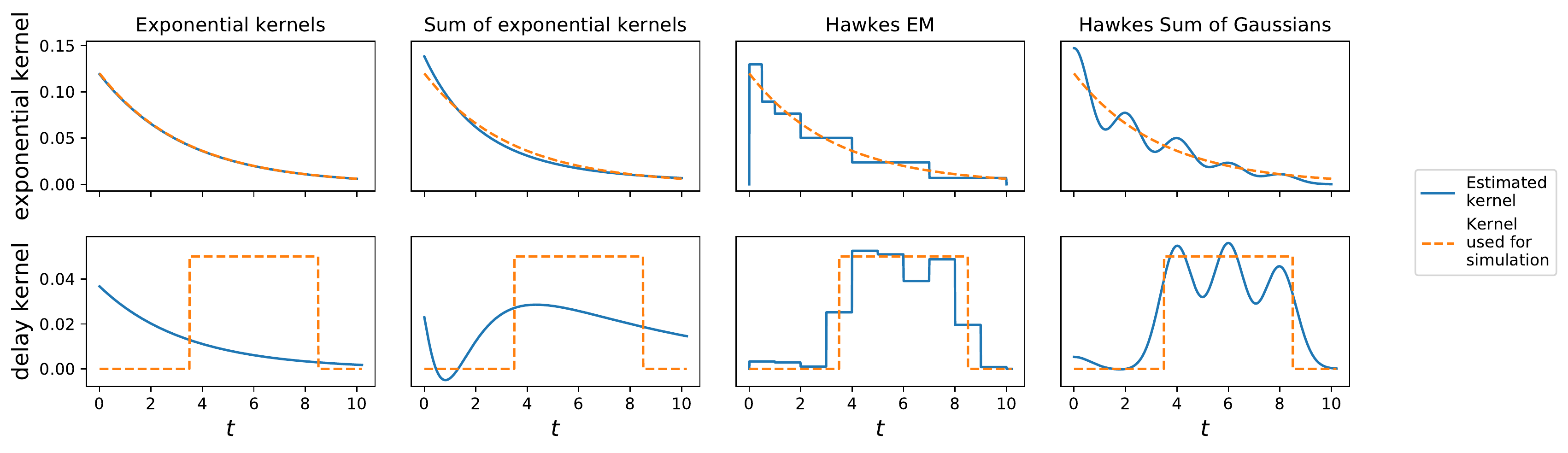}
  \caption{\small Illustration of different kernels shapes and estimations obtained by \texttt{tick} on two 1D simulated Hawkes processes with intensity kernels displayed with dashed orange lines.}
  \label{fig:hawkes_fit_kernel}
\end{figure}

\begin{figure}
\begin{minipage}{.6\textwidth}
  \includegraphics[height=0.14\textheight]{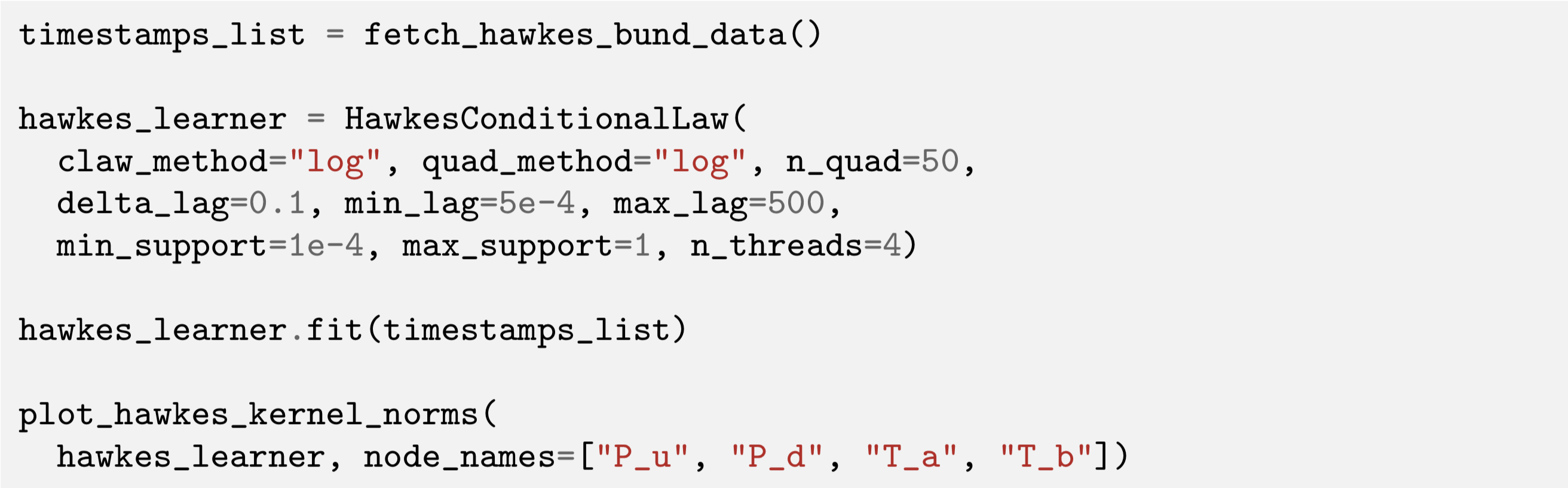}
\end{minipage}%
\begin{minipage}{.4\textwidth}
  \centering
  \includegraphics[width=0.2\textheight]{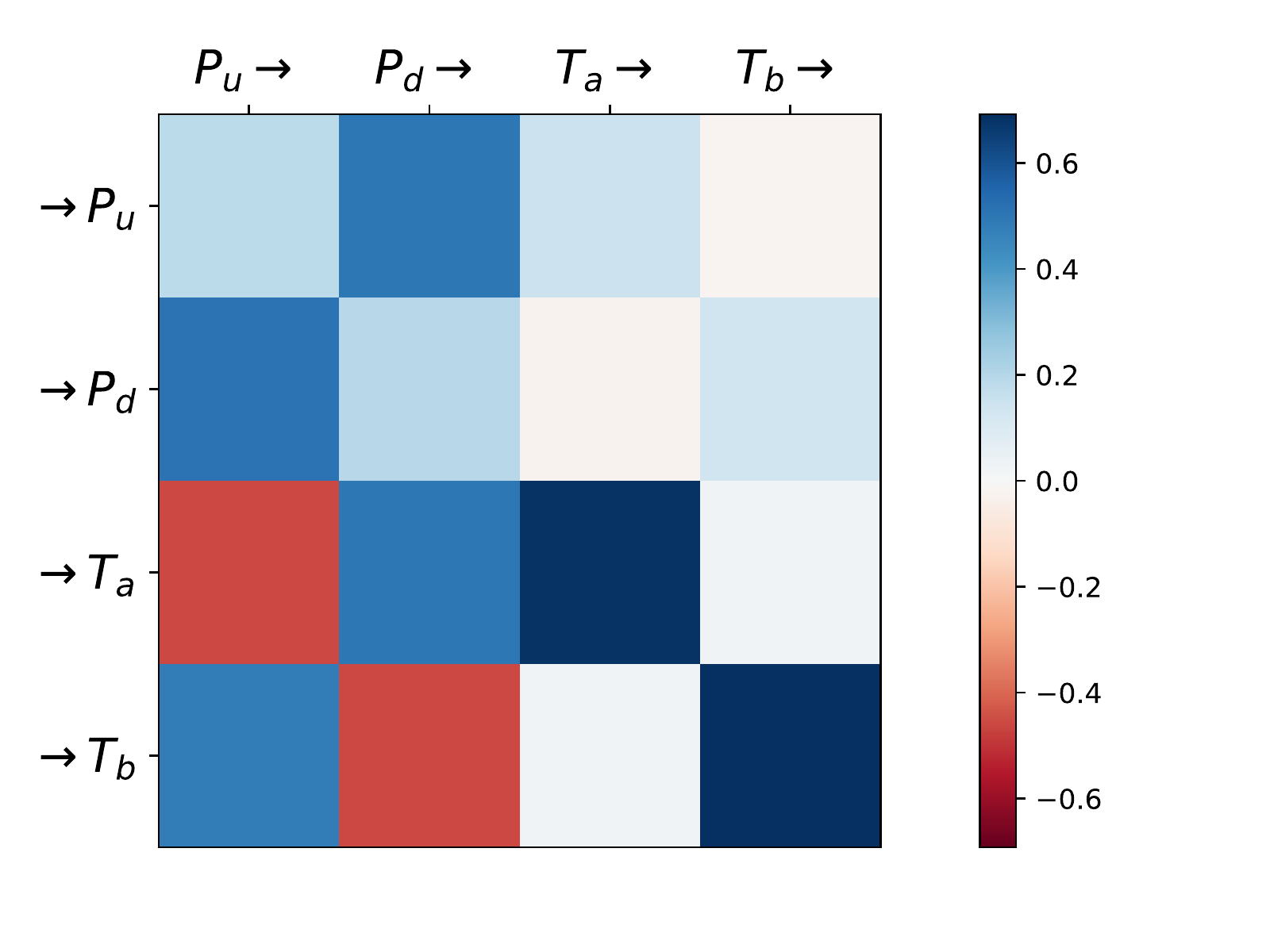}
\end{minipage}%
\caption{\small Kernels norms of a Hawkes process fitted on high-frequency financial data from the Bund market \citep{bacry2014estimation} where $P_u$ (resp. $P_d$) counts the number of upward (resp. downward) mid-price moves and $T_a$ (resp. $T_b$) counts the number of market orders at the ask (resp. bid) that do not move the price.
}
\label{fig:bund}
\end{figure}

\begin{figure}
  \centering
  \includegraphics[height=0.2\textheight]{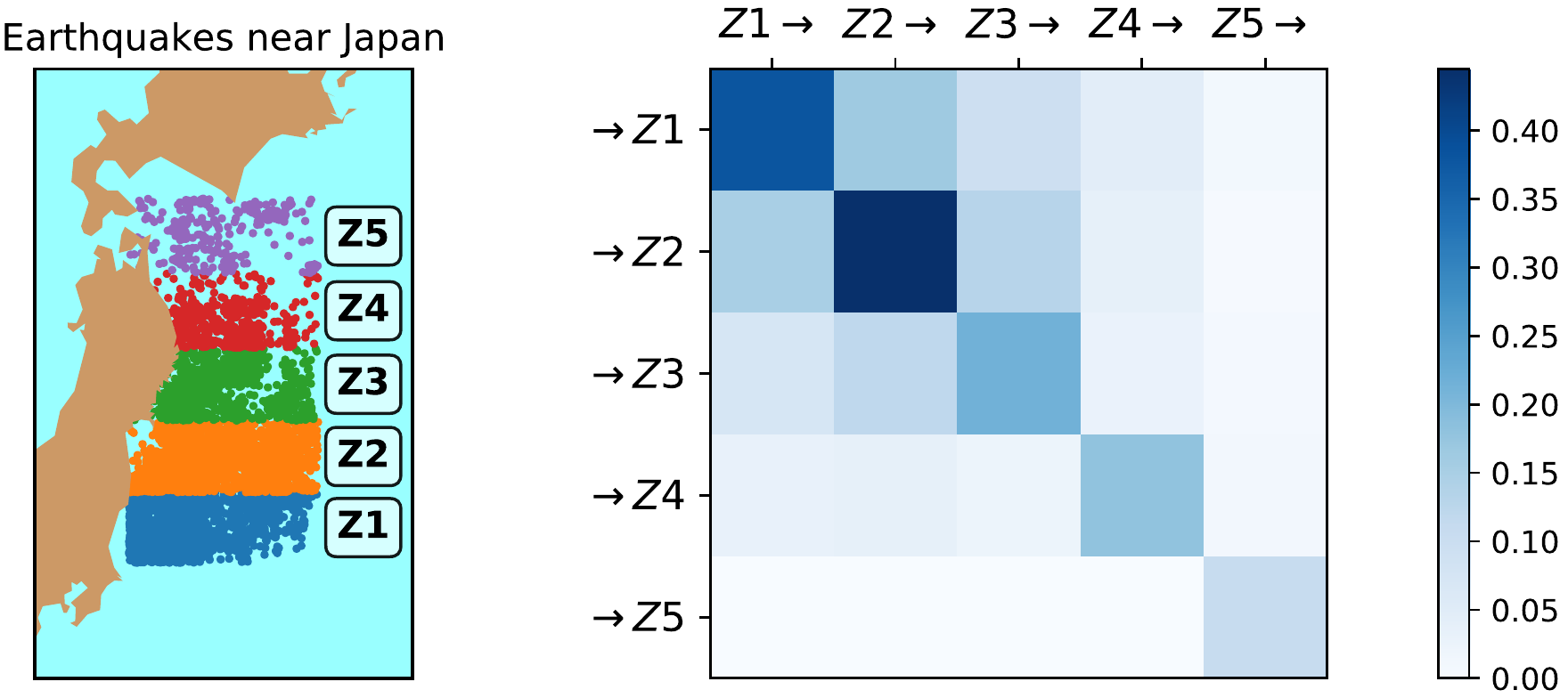}
  \caption{\small Analysis with Hawkes processes of earthquake propagation with a dataset from \cite{ogata1988statistical}. On the left we can see where earthquakes have occurred and on the right the propagation matrix, i.e. how likely a earthquake in a given zone will trigger an aftershock in another zone. We can observe than zone 1, 2 and 3 are tightly linked while zone 4 and 5 are more self-excited.}
  \label{fig:earthquakes}
\end{figure}

Distributing a comprehensive open source library for Hawkes processes is one of the primary aims of the \tick library: it provides many non-parametric and parametric estimation algorithms as well as simulation tools for many kernel types, that are listed in Table~\ref{tab:hawkes_estimators}.
\begin{figure}
  \centering
  \small
  \newcolumntype{C}{>{\centering\arraybackslash}X}%
  \begin{tabularx}{\linewidth}{ C C }
    Non Parametric & Parametric \\
    \hline \hline
    EM \citep{lewis2011nonparametric} & Single exponential kernel  \\
    Basis kernels \citep{zhou2013learning} & Sum of exponentials kernels \\
    Wiener-Hopf \citep{bacry2014second} & Sum of gaussians kernels \citep{xu2016learning} \\
    NPHC \citep{achab2017uncovering}  & ADM4 \citep{zhou2013learning}
  \end{tabularx}
  \captionof{table}{\small Hawkes estimation algorithms implemented in \tick{}}
  \label{tab:hawkes_estimators}
\end{figure}
This diversity of algorithms is illustrated in Figure~\ref{fig:hawkes_fit_kernel} (with the associated Python code) in which we show how two kernels of different shapes are estimated by four different algorithms. A first use case for modeling high-frequency financial data is given in Figure~\ref{fig:bund}, while 
a second use-case about propagation analysis of earthquake aftershocks can be found in Figure~\ref{fig:earthquakes}.

\section{Benchmarks}

\begin{figure}
  \centering
  \includegraphics[height=0.21\textheight]{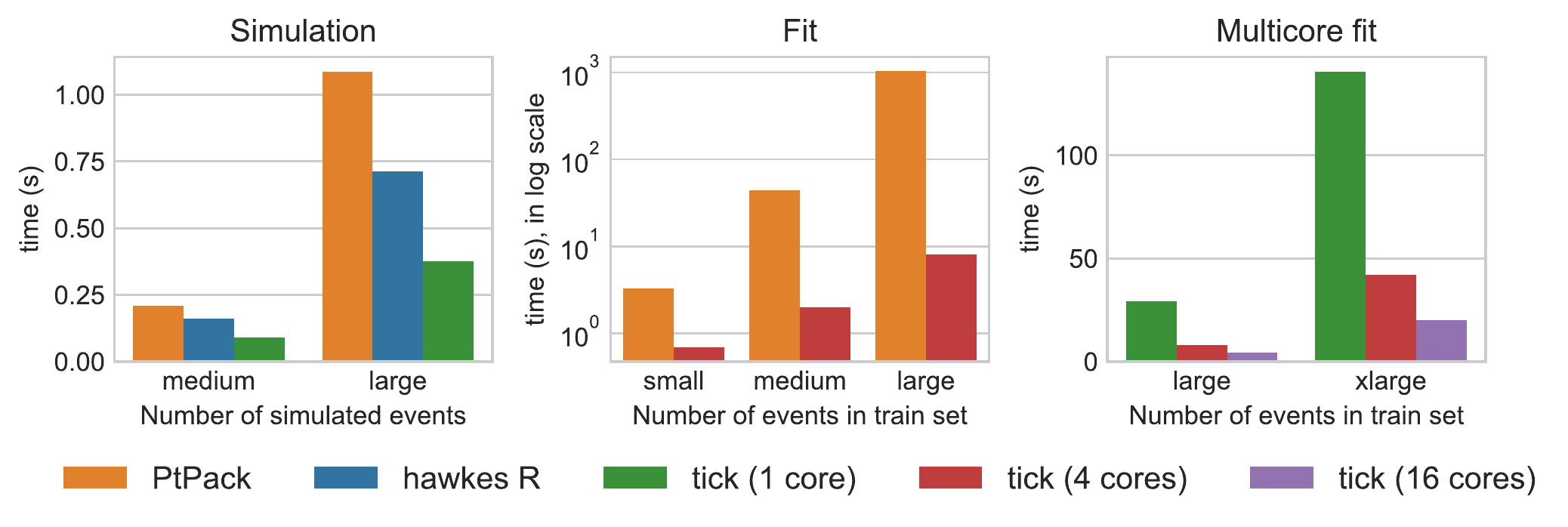}
  \caption{\small Computational timings of \tick{} versus \ptpack{} and \hawkesR{}. \tick strongly outperforms both libraries for simulation and fitting (note that fit graph is in log-scale). 
  Third figure shows that \tick benefits from multi-core environments to speed up computations.}
  \label{fig:benchmark_hawkes}
\end{figure}

We perform benchmark tests for both simulation and estimation of Hawkes processes (with exponential kernels) using \tick, \hawkesR{} (where only simulation is available) and \ptpack{}, on respectively 2, 4 and 16 cores.
In Figure~\ref{fig:benchmark_hawkes} we compare computational times for simulation and fitting of Hawkes processes.
The model fits are compared on simulated 16-dimensional Hawkes processes, with an increasing number of events: small=$5\times 10^4$, medium=$2 \times 10^5$, large=$10^6$, xlarge$=5 \times 10^7$.
We observe on this experiment that \tick outperforms by several orders of magnitudes both \hawkesR{} and \ptpack{}, in particular for large datasets.
Benchmarks against \sklearn{} for logistic regression are also provided in Figure~\ref{fig:tick_vs_scikit} from the supplementary material.

\section*{Acknowledgments}

We would like to acknowledge support for this project from the Datascience Initiative of \'Ecole polytechnique and Intel\textsuperscript{\tiny\textregistered} for supporting \tick development.

\newpage

\vskip 0.2in

\newpage
\renewcommand\appendixtocname{Appendix}
\renewcommand\appendixpagename{Appendix}
\begin{appendices}



\section{Speed comparison} 
\label{sec:speed_comparison}

\begin{figure}
  \centering
  \includegraphics[width=\linewidth]{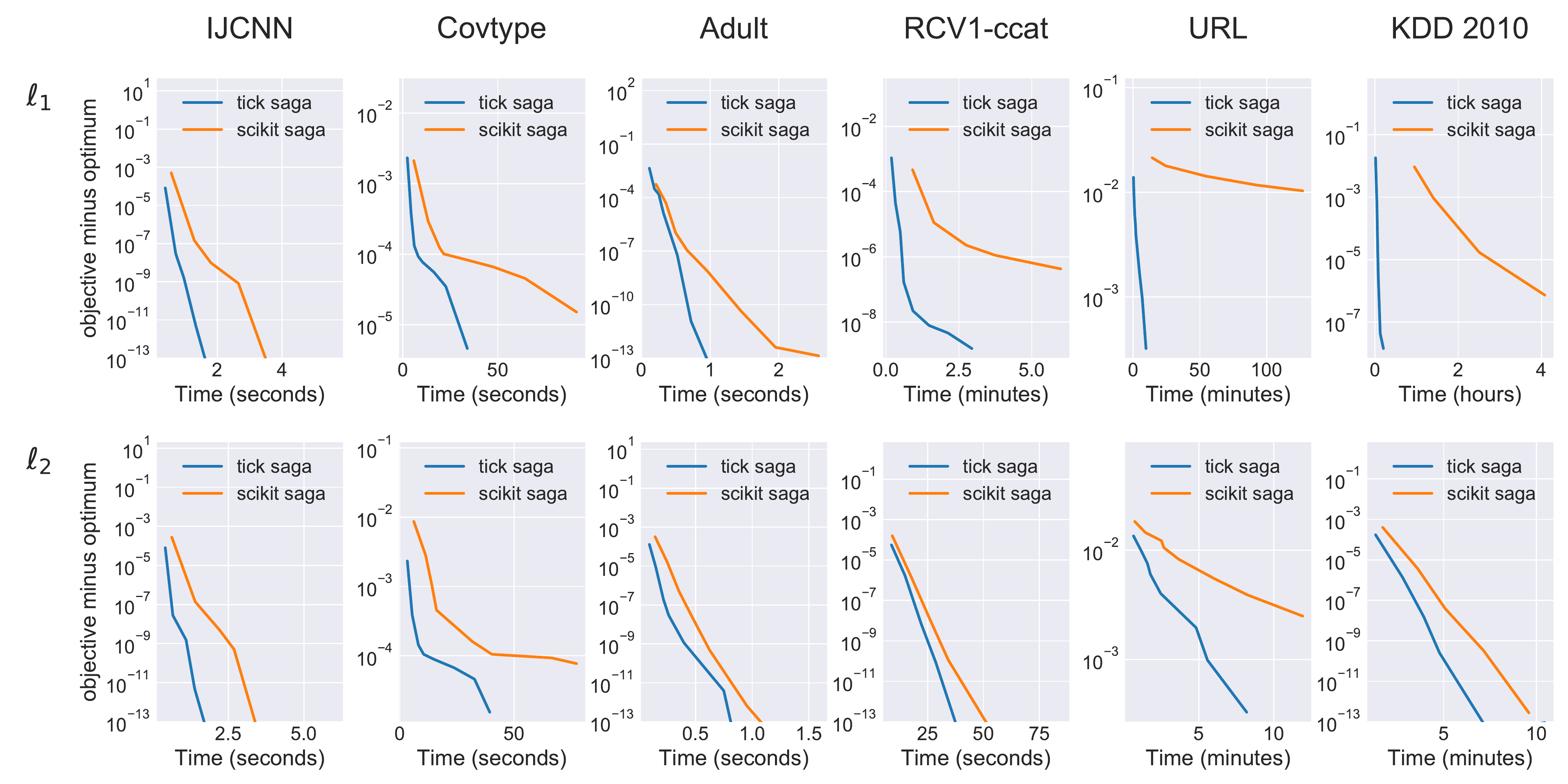}
  \caption{\small Speed comparison with \sklearn{} library. 
  These plots display time needed to achieve a given precision for logistic regression with $\ell_1$ and $\ell_2$ penalizations on commonly used datasets.
  In both cases we use SAGA solver as the two libraries provide it.}
  \label{fig:tick_vs_scikit}
\end{figure}

We compare fitting results for binary logistic regression with \tick and \sklearn. These experiments are run on commonly used datasets described in Table~\ref{tab:datasets}. Note that Covtype has been standardized, hence the first two datasets IJCNN and Covtype are dense and the last four datasets are sparse. Two types of penalization have been tested: $\ell_1$ (Lasso) and $\ell_2$ (Ridge). In both cases the regularization parameter $\lambda$ has been set to $1 / n$ where $n$ is the number of samples and we have left the default step-size for both libraries. Results are given in Figure~\ref{fig:tick_vs_scikit}. Overall, \tick is slightly faster because it makes faster iterations: both libraries reach the same objective after each pass over the data but \tick performs these computations faster. Also, $\ell_1$ penalization in high dimension is difficult for \sklearn{} (see URL and KDD 2010) whereas \tick handles it without any additional problem.

\begin{figure}
  \centering
  \small
  \newcolumntype{C}{>{\centering\arraybackslash}X}%
  \begin{tabularx}{\linewidth}{ C C C C}
    dataset & \# samples & \# features & density \\
    \hline \hline
    IJCNN & 141,691 & 22 & 100 \% \\
    Covtype & 581,012 & 54 & 100 \% \\
    Adult & 32,561 & 123 & 11.3 \% \\
    RCV1-ccat & 804,414 & 47,236 & 0.0016 \% \\
    URL & 2,396,130 & 3,231,961 & 0.000036 \% \\
    KDD 2010 & 19,264,097 & 1,163,024 & 0.00078 \%
  \end{tabularx}

  \captionof{table}{Datasets used to perform binary logistic regression.}
  \label{tab:datasets}
\end{figure}

\section{Package structure} 
\label{sec:package_structure}

\begin{figure}
  \centering
  \includegraphics[width=\linewidth]{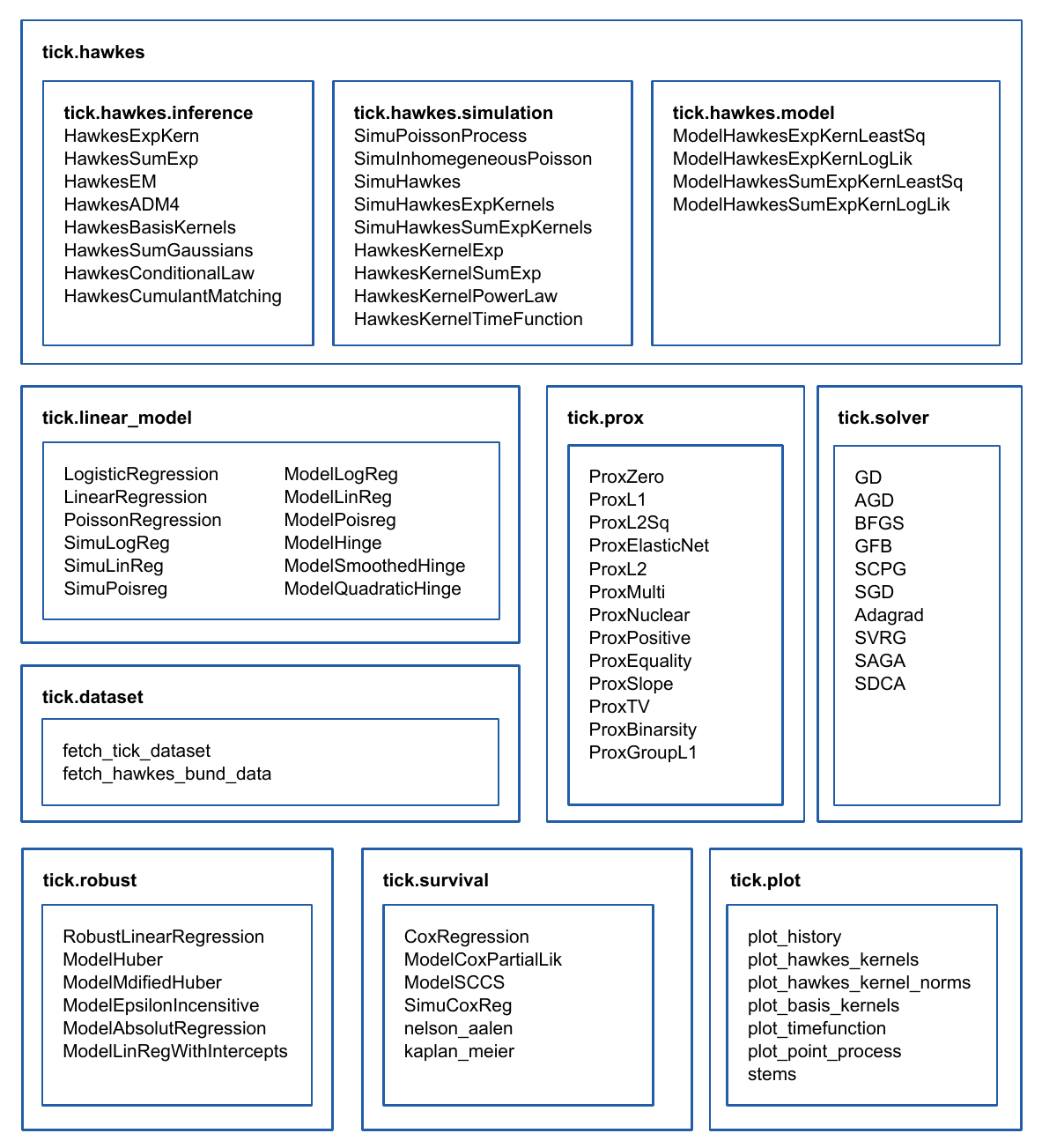}
  \caption{Structure of \tick package}
  \label{fig:tick_structure}
\end{figure}

The package structure is detailed in Figure~\ref{fig:tick_structure}. We retrieve all the following modules:
\begin{itemize}
  \item \texttt{tick.hawkes} : Inference and simulation of Hawkes processes, with both parametric and non-parametric estimation techniques and flexible tools for simulation. It is split in three submodules:
  \texttt{tick.hawkes.inference}, \texttt{tick.hawkes.simulation}, \texttt{tick.hawkes.model}.

  \item \texttt{tick.linear\_model} : Inference and simulation of linear models, including among others linear, logistic and Poisson regression, with a large set of penalization techniques and solvers. 

  \item \texttt{tick.robust} : Tools for robust inference. It features tools for outliers detection and models such as Huber regression, among others robust losses. 

  \item \texttt{tick.survival} :  Inference and simulation for survival analysis, including Cox regression with several penalizations. 

  \item \texttt{tick.prox} : Proximal operators for penalization of models weights. Such an operator can be used with (almost) any model and any solver. 

  \item \texttt{tick.solver} : A module that provides a bunch of state-of-the-art optimization algorithms, including both batch and stochastic solvers 

  \item \texttt{tick.dataset} : Provides easy access to datasets used as benchmarks in tick. 

  \item \texttt{tick.plot} : Some plotting utilities used in tick, such as plots for point processes and solver convergence. 
\end{itemize}

\end{appendices}

\end{document}